\pdfoutput=1

\documentclass[11pt]{article}

\usepackage{naacl2021}

\usepackage{times}
\usepackage{latexsym}
\usepackage{graphicx}
\usepackage{framed}
\usepackage[normalem]{ulem}
\usepackage{booktabs}
\usepackage{array}
\usepackage{xspace}
\usepackage{comment}
\usepackage{booktabs}
\usepackage{enumitem}
\usepackage{multirow}
\usepackage{xcolor}
\usepackage{makecell}
\usepackage{subfig}
\usepackage{amsmath}
\usepackage{amsfonts}

\newenvironment{itemize*}%
 {\leftmargini=10pt\begin{itemize}%
  \setlength{\itemsep}{0pt}%
  \setlength{\parskip}{0pt}%
  }%
 {\end{itemize}}
\newenvironment{enumerate*}%
 {\begin{enumerate}%
  \setlength{\itemsep}{0pt}%
  \setlength{\parskip}{0pt}}%
 {\end{enumerate}}

\usepackage[T1]{fontenc}

\usepackage[utf8]{inputenc}

\usepackage{microtype}

\definecolor{myblue}{rgb}{0.9, 0.1, 0.94}
\definecolor{mygreen}{rgb}{0.64, 0.56, 0.88}
\definecolor{myyellow}{rgb}{0.98, 0.94, 0.75}
\definecolor{mygreen}{rgb}{0.68, 0.9, 0.6}

\definecolor{myorange}{rgb}{1.0, 0.49, 0.0}

\title{\textsc{ExplainaBoard}:\\ An Explainable Leaderboard for NLP}

\author{Pengfei Liu$^{1\dag}$, \;\; Jinlan Fu$^2$, \;\; Yang Xiao$^2$, \;\; Weizhe Yuan$^1$, \;\; Shuaichen Chang$^3$,  \\ \bf Junqi Dai$^2$, \;\; Yixin Liu$^1$, \;\; Zihuiwen Ye$^1$, \;\; Zi-Yi Dou$^1$, \;\; Graham Neubig$^{1\ddag}$   \\ 
  $^1$Carnegie Mellon University,  $^2$Fudan University, $^3$The Ohio State University,    \\
  $^\dag$\texttt{pliu3@cs.cmu.edu,} $^\ddag$\texttt{gneubig@cs.cmu.edu} \\
  }

\begin{document}
\maketitle
\begin{abstract}

With the rapid development of NLP research, leaderboards
have emerged as one tool to track the performance of various systems on various NLP tasks.
They are effective in this goal to some extent, but generally present a rather simplistic one-dimensional view of the submitted systems, communicated only through holistic accuracy numbers.
In this paper, we present a new conceptualization and implementation of NLP evaluation: the \textsc{ExplainaBoard}, which in addition to inheriting the functionality of the standard leaderboard, also allows researchers to (i) diagnose strengths and weaknesses of a single system (e.g.~what is the best-performing system bad at?) (ii) interpret relationships between multiple systems. (e.g.~where does system \texttt{A} outperform system \texttt{B}? What if we combine systems \texttt{A}, \texttt{B} and \texttt{C}?) and (iii) examine prediction results closely (e.g.~what are common errors made by multiple systems or in what contexts do particular errors occur?). 
So far, \textsc{ExplainaBoard} covers more than 400 systems, 50 datasets, 40 languages, and 12 tasks.%
\footnote{\textsc{ExplainaBoard} keeps updated and is recently upgraded by supporting  (1) \textbf{multilingual multi-task benchmark}, (2) \textbf{meta evaluation}~\cite{yuan2021bartscore} and (3) more complicated task: \textbf{machine translation}, which reviewers also suggested.} 
We not only released an online \textbf{platform} at the website
\footnote{\url{http://explainaboard.nlpedia.ai/}} but also make our evaluation tool an \textbf{API} with \textit{MIT Licence} at Github
\footnote{\url{https://github.com/neulab/explainaBoard}} and PyPi
\footnote{\url{https://pypi.org/project/interpret-eval/}} that allows users to conveniently assess their models offline.
We additionally release all \textbf{output files} from systems that we have run or collected to motivate ``output-driven'' research in the future.

\end{abstract}

\section{Introduction}

Natural language processing (NLP) research has been and is making astounding strides forward.
This is true both for classical tasks such as machine translation \cite{DBLP:conf/nips/SutskeverVL14,45610}, as well as for new tasks \cite{DBLP:conf/nips/LuYBP16,rajpurkar-etal-2016-squad}, domains \cite{beltagy-etal-2019-scibert}, and languages \cite{DBLP:conf/nips/ConneauL19}. 
 One way this progress is quantified is through \emph{leaderboards}, which report and update performance numbers of state-of-the-art systems on one or more tasks. Some prototypical leaderboards include GLUE and SuperGLUE  for natural language understanding \cite{wang-etal-2018-glue,wang2019superglue}, XTREME and XGLUE \cite{hu2020xtreme,liang2020xglue} for multilingual understanding, the WMT shared tasks \cite{barrault-etal-2020-findings} for machine translation,
and GEM  and GENIE  for natural language generation \cite{gehrmann2021gem,khashabi2021genie}, among many others.


\begin{figure}%
\centering
\includegraphics[width=0.98\linewidth]{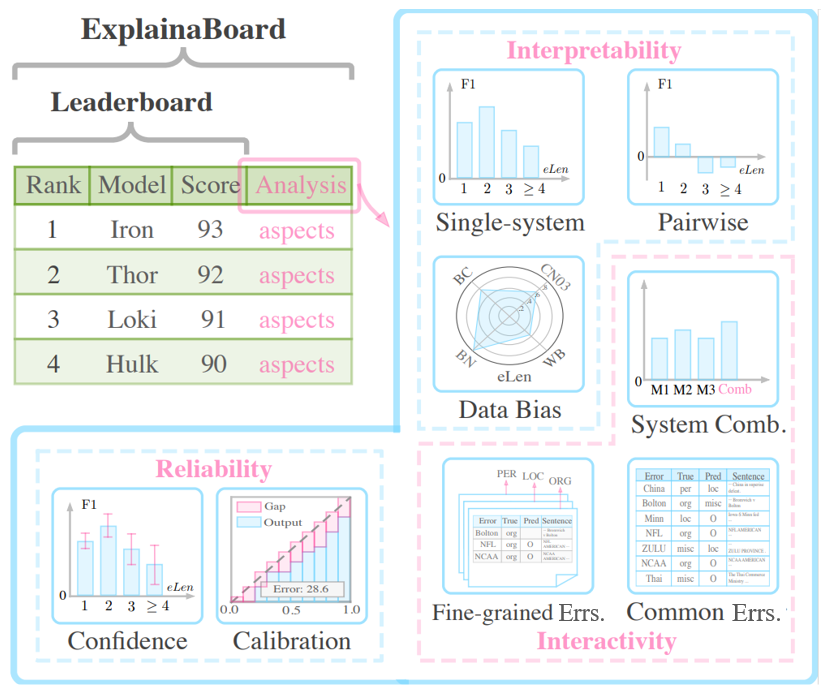}

\caption{ Illustration of the \textsc{ExplainaBoard} concept.
Compared to vanilla leaderboards, \textsc{ExplainaBoard}  allows users to perform \textit{interpretable} (single-system , pairwise analysis, data bias), \textit{interactive} (system combination, fine-grained/common error analysis), and \textit{reliable} analysis (confidence interval, calibration) on systems in which they are interested. ``\texttt{Comb.}'' denotes ``combination'' and ``\texttt{Errs}'' represents ``errors''. ``\texttt{PER}, \texttt{LOC}, \texttt{ORG}'' refer to different labels.
                            } 
\label{fig:intro}
\end{figure}

These leaderboards serve an important purpose: they provide a standardized evaluation setup, often on multiple tasks, that eases reproducible model comparison across organizations. However, at the same time, due to the prestige imbued by a top, or high, place on a leaderboard they also can result in a singular focus on raising evaluation numbers at the cost of deeper scientific understanding of model properties \citep{ethayarajh-jurafsky-2020-utility}. In particular, we argue that, among others, the following are three major limitations of the existing leaderboard paradigm:

\begin{itemize*}
    \item \textbf{Interpretability:} Most existing leaderboards commonly use a single number to summarize system performance holistically. This is conducive to system ranking but at the same time, the results are opaque, making the strengths and weaknesses of systems less interpretable.
    \item \textbf{Interactivity:} Existing leaderboards are static and non-interactive, which limits the ability of users to dig deeper into the results. Thus, (1) they usually do not flexibly support more complex evaluation settings (e.g.~multi-dataset, multi-metric, multi-language) (2) users may miss opportunities to understand the relationships between different systems. For example, where does model A outperform model B? Would the performance be further improved if we combine the Top-$3$ models?
    \item \textbf{Reliability:} Given the increasing role that leaderboards have taken in guiding NLP research, it is important that information expressed in them is reliable, especially on datasets with small sample sizes, but most current leaderboards do not give an idea of the reliability of system rankings.
\end{itemize*}

\renewcommand\tabcolsep{3pt}
\renewcommand\arraystretch{1.6}  
\begin{table*}[htb]
\vspace{-3mm}
\small
  \centering 
    \begin{tabular}{ccccc}
    \toprule
    \textbf{Aspect} & \textbf{Functionality} & \textbf{Input} & \multicolumn{2}{c}{\textbf{Output}} \\
    \midrule
    \multirow{9}[2]{*}{Interpretability} & \multirowcell{3}{Single-system \\ Analysis} & \multirow{3}[1]{*}{One model} & \multirow{3}[1]{*}{\includegraphics[scale=0.55]{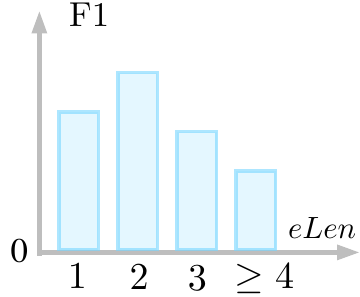}} &
    \multirow{3}[1]{*}{\makecell[{{p{6cm}}}]{\textbf{Performance Histogram}: the input model is good at dealing with short entities, while achieving lower performance on long entities.}} 
                        \\ \\ \\
                        
    \cmidrule(lr){2-5}
     & \multirowcell{3}{Pairwise \\ Analysis} & \multirowcell{3}{Two models \\(M1,M2)} & \multirow{3}[1]{*}{\includegraphics[scale=0.55]{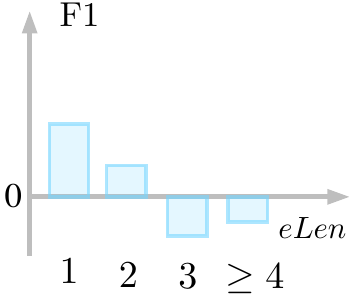}} & \multirow{3}[1]{*}{\makecell[{{p{6cm}}}]{\textbf{Performance Gap Histogram} (M1$-$M2): M1 is better at dealing with short entities, while M2 is better at dealing with long entities.}} \\ \\ \\
          \cmidrule(lr){2-5}
          & \multirowcell{3}{Data Bias \\ Analysis} & \multirow{3}[1]{*}{Multi-dataset} & \multirow{3}[1]{*}{\includegraphics[scale=0.7]{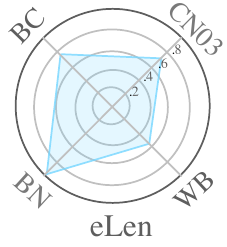}} & \multirow{3}[1]{*}{\makecell[{{p{6cm}}}]{\textbf{Data Bias Chart}: For the \texttt{entity length} attribute, the average entity length (We average the length of all test entities on a given data set.) of these datasets order by descending is  
          BN$\textgreater$
          BC$\textgreater$
          CN03$\textgreater$
          WB.
          }}\\ \\ \\
          \midrule
          \multirow{6}[2]{*}{Interactivity} & \multirowcell{3}{Fine-grained \\ Error Analysis}& \multirowcell{3}{Single- or \\ Pairwise-system \\ diagnostic results} & \multirow{3}[1]{*}{\includegraphics[scale=0.32]{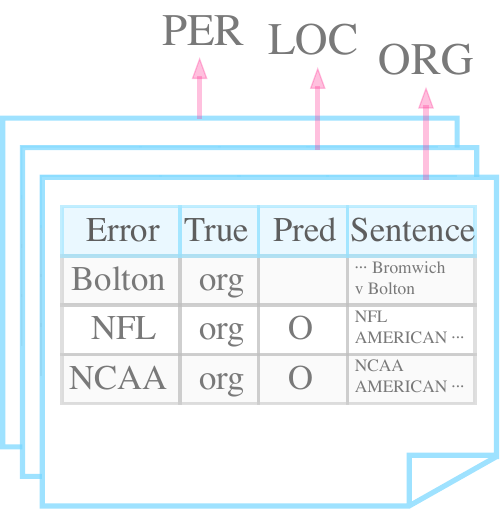}} & \multirow{3}[1]{*}{\makecell[{{p{6cm}}}]{\textbf{Error Table}: Error analysis allows the user to print out the entities that are incorrectly predicted by the given model, as well as the true label of the entity, the mispredicted label, and the sentence where the entity is located.}}\\
         \\ \\ 
          \cmidrule(lr){2-5}
          & \multirowcell{3}{System \\ Combination }& \multirowcell{3}{Multi-models \\(M1,M2,M3)} & \multirow{3}[1]{*}{\includegraphics[scale=0.55]{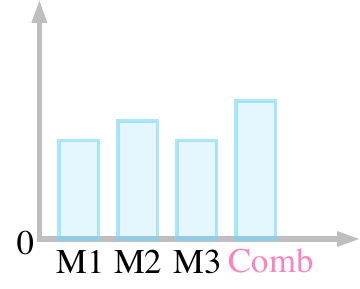}} & \multirow{3}[1]{*}{\makecell[{{p{6cm}}}]{\textbf{Ensemble Chart}: The combined result of model M1, M2, and M3 is shown by the histogram with x-label value \texttt{comb}. The combined result is better than the single models.}}\\
         \\ \\
      
    \midrule
    \multirow{6}[2]{*}{Reliability} & \multirow{3}[1]{*}{Confidence} & \multirow{3}[1]{*}{One model} & \multirow{3}[1]{*}{\includegraphics[scale=0.55]{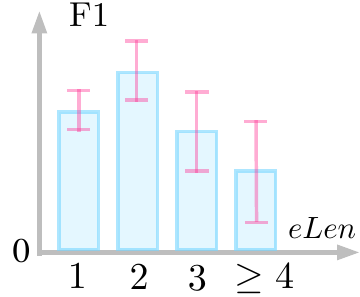}} & \multirow{3}[1]{*}{\makecell[{{p{6cm}}}]{\textbf{Error Bars}: the error bars represent 95$\%$ confidence intervals of the performance on the specific bucket.}}\\
         \\ \\
          \cmidrule(lr){2-5}
          & \multirow{3}[1]{*}{Calibration} & \multirow{3}[1]{*}{One model} & \multirow{3}[1]{*}{\includegraphics[scale=0.72]{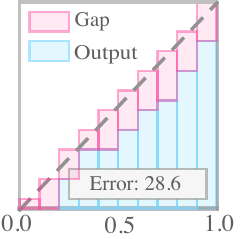}} &
          \multirow{3}[1]{*}{\makecell[{{p{6cm}}}]{\textbf{Reliability Diagram}: Confidence histograms (red) and reliability diagrams (blue). that indicate the accuracy of model probability estimates}}
          \\
         \\ \\
    \bottomrule
    \end{tabular}%
      \caption{A graphical breakdown of the functionality of \textsc{ExplainaBoard}, with examples from an NER task. \vspace{-3mm}}
  \label{tab:functionalities}%
\end{table*}%

In this paper, we describe \textsc{ExplainaBoard} (see Fig.\ref{fig:intro}), a software package and hosted leaderboard that satisfies all of the above desiderata. It also serves as a prototype implementation of some desirable features that may be included in future leaderboards, even independent of the provided software itself.
We have deployed \textsc{ExplainaBoard} for 9 different tasks and 41 different datasets, and demonstrate how it can be easily adapted to new tasks of interest.

We expect that \textsc{ExplainaBoard} will benefit different steps of the research process:

\noindent 
(i) \textbf{System Developement}:  \textsc{ExplainaBoard} provides more detailed information regarding the submitted systems (e.g.~fine-grained results, confidence intervals), allowing system developers to diagnose successes and failures of their own systems, or compare their systems with baselines and understand where  improvements of their proposed methods come from.
This better understanding can lead to more efficient and effective system improvements.
Additionally, \textsc{ExplainaBoard} can help system developers uncover their systems' advantages over others, even when these systems have not achieved state-of-the-art performance holistically.
(ii) \textbf{Leaderboard Organization}: The \textsc{ExplainaBoard} software both provides a ready-made platform for easy leaderboard development over different NLP tasks, and helps upgrade traditional leaderboards to allow for more fine-grained analysis. For example, we have already established an ExplainaBoard~\footnote{\url{http://explainaboard.nlpedia.ai/leaderboard/xtreme/}} for the existing XTREME benchmark.\footnote{\url{https://sites.research.google/xtreme/}}
(iii) \textbf{Broad Analysis and Understanding}: 
Because \textsc{ExplainaBoard} encourages system developers to provide their system outputs in an easy-to-analyze format, these will also help researchers, particularly those just starting in a particular NLP sub-field, get a broad sense of what current state-of-the-art models can and cannot do.
This not only helps them quickly track the progress of different areas, but also can allow them to understand the relative advantages of diverse systems, suggesting insightful ideas for what's left and what's next.\footnote{Since the first release of \textsc{ExplainaBoard}, we have received invitations from multiple companies, startups, and researchers to collaborate, and we are working together to make it better for the community.}

\section{ExplainaBoard}

As stated above, \textsc{ExplainaBoard} extends existing leaderboards, improving their interpretability, interactivity, and reliability.
It does so by providing a number of functionalities that are applicable to a wide variety of NLP tasks (as illustrated in Tab.~\ref{tab:functionalities}).
Many of these functionalities are grounded in existing research on evaluation and fine-grained diagnostics.

\subsection{Interpretability}

\emph{Interpretable evaluation} \cite{popovic-ney-2011-towards,stymne-2011-blast,neubig-etal-2019-compare,fu-etal-2020-interpretable}, 
is a research area that considers methods that break down the holistic performance of each system into different interpretable groups.
For example, in a Named Entity Recognition (NER) task, we may examine the accuracy along different dimensions of a concerned entity (such as ``entity frequency,'' telling us how well the model does on entities that appear in the training data a certain number of times) or sentences (such as ``sentence length,'' telling us how well the model does on entities that appear in longer or shorter sentences) \cite{fu-etal-2020-interpretable}.
This makes it possible to understand where models do well and poorly, leading to further insights beyond those that can be gleaned by holistic evaluation numbers.
Applying this to a new task involves the following steps:
(i) \textit{Attribute definition}: define attributes by which we can partition the test set into different groups.
\noindent (ii) \textit{Bucketing}: partition into different buckets based on defined attributes and calculate performance w.r.t each bucket.

Generally, previous work on interpretable evaluation has been performed over single tasks, while \textsc{ExplainaBoard} allows for comprehensive evaluation of different types of tasks in a single software package.
We concretely show several ways interpretable evaluation can be defined within \textsc{ExplainaBoard} below:

\paragraph{F1\footnote{``F'' represents ``Functionality''.}: Single-system Analysis:  What is a system good or bad at?}
For an individual system as input, generate a \emph{performance histogram} that highlights the buckets where it performs well or poorly.
For example, in Tab.~\ref{tab:functionalities} we demonstrate an example from NER where the input system does worse in dealing with longer entities ($\text{eLen} \geq 4$).

\paragraph{F2: Pairwise Analysis: Where is one system better (worse) than another?}
Given a pair of systems, interpret where the performance gap occurs.
Researchers could flexibly choose two systems they are interested in (e.g.~selecting two rows from the leaderboard), and \textsc{ExplainaBoard} will output a \textit{performance gap histogram} to describe how the \textbf{performance differences} change over different buckets of different attributes. Tab.~\ref{tab:functionalities} demonstrates how we can see one system is better than the other at longer or shorter entities.

\paragraph{F3: Data Bias Analysis: What are the characteristics of different evaluated datasets?}
The defined attributes do not only help us interpret system performance, but also make it possible for users to take a closer look at characteristics of diverse datasets.
For example, from Fig.~\ref{tab:functionalities} shows an example of analyzing differences in average entity length across several datasets.

\subsection{Interactivity}
\textsc{ExplainaBoard} also allows users to dig deeper, interacting with the results in more complex ways.

\paragraph{F4: Fine-grained Error Analysis: What are common mistakes that most systems make and where do they occur?}
\textsc{ExplainaBoard} provides flexible fine-grained error analyses based on the above-described performance evaluation:
\begin{enumerate*}
    \item Users can choose multiple systems and see their \textit{common error cases}, which can be useful to identify \textit{challenging samples} or even \textit{annotation errors}.
    \item In single-system analysis, users can choose particular buckets in the performance histogram\footnote{Each bin of the performance histogram is clickable, returning an error case table.} and see corresponding error samples in that bucket (e.g.~which long entities does the current system mispredict?).
    \item In pairwise analysis, users can select a bucket, and the unique errors (e.g.~system A succeeds while B fails and vice versa) of two models will be displayed.
\end{enumerate*}

\paragraph{F5: System Combination: Is there potential complementarity between different systems?}
System combination \cite{DBLP:conf/ijcai/TingG97,gonzalez-rubio-etal-2011-minimum, duh2011generalized} is a technique to improve performance by combining the output from multiple existing systems.
In \textsc{ExplainaBoard},
users can choose multiple systems and obtain combined results calculated by voting over multiple base systems.\footnote{With the system combination button of Explainaboard, we observed the-state-of-the art performance of some tasks (e.g., NER, Chunking) can be further improved.}
In practice, for NER task, we use the recently proposed \textsc{SpanNer} \cite{jinspanner2021} as a combiner, and for text summarization we employed \textsc{Refactor}, a state-of-the-art ensemble approach \cite{liu2020refactor}.
Regarding the other tasks, we adopt the majority voting method for system combination.

\subsection{Reliability}
The experimental conclusions obtained from the evaluation metrics are not necessarily statistically reliable, especially when the experimental results can be affected by many factors.
\textsc{ExplainaBoard} also makes a step towards more reliable interpretable evaluation.

\paragraph{F6: Confidence Analysis: To what extent can we trust the results of our system?}
\textsc{ExplainaBoard} can perform confidence analysis over both holistic and fine-grained performance metrics.
As shown in Tab.~\ref{tab:functionalities}, for each bucket, there is an error bar whose width reflects how reliable the performance value is.
We claim this is an important feature for fine-grained analysis since the numbers of test samples in each bucket are imbalanced, and with the confidence interval, one could know how much uncertainty there is.
In practice, we use bootstrapping method \cite{efron1992bootstrap,ye21eacl} to calculate the confidence interval.

\paragraph{F7: Calibration Analysis: How well is the confidence of prediction calibrated with its correctness?}
One commonly-cited issue with modern neural predictors is that their probability estimates are not accurate (i.e.~they are poorly \emph{calibrated}), often being over-confident in the correctness of their predictions \cite{DBLP:conf/icml/GuoPSW17}.
We also incorporate this feature into \textsc{ExplainaBoard}, allowing users to evaluate how well-calibrated their systems of interest are.

\section{Tasks, Datasets and Systems}

We have already added to \textsc{ExplainaBoard} 12 NLP tasks, 50 datasets, and 400 models,\footnote{265 of these models are implemented by us, as unfortunately it is currently not standard in NLP to release the system outputs that \textsc{ExplainaBoard} needs.} which cover many or most of top-scoring systems on these tasks.
We briefly describe them below, and show high-level statistics in  Tab.~\ref{tab:task_data_model}.

\paragraph{Text Classification}
Prediction of one or multiple pre-defined label(s) for a given input text.
The current interface includes datasets for sentiment classification~\cite{pang-etal-2002-thumbs}, topic identification~\cite{wang-manning-2012-baselines}, and intention detection~\cite{chen-etal-2013-identifying}.

\paragraph{Text-Span Classification}
Prediction of a pre-defined class from the input of a text and a span, such as aspect-based sentiment classification task \cite{pappas-popescu-belis-2014-explaining}.
We collect top-perform system outputs from \cite{dai2021does}.

\paragraph{Text Pair Classification}
Prediction of a class given two texts, such as the natural language inference task \cite{bowman-etal-2015-large}.

\paragraph{Sequence Labeling}
Prediction of a label for each token in a sequence. The \textsc{ExplainaBoard} currently includes four concrete tasks: named entity recognition \cite{tjong-kim-sang-de-meulder-2003-introduction}, part-of-speech tagging \cite{toutanova-etal-2003-feature}, text chunking \cite{ando-zhang-2005-high}, and Chinese word segmentation \cite{chen-etal-2015-long}.

\paragraph{Structure Prediction}
Prediction of a syntactic or semantic structure from text, where \textsc{ExplainaBoard} currently covers semantic parsing tasks~\cite{berant-etal-2013-semantic,yu-etal-2018-spider}.

\paragraph{Text Generation}
\textsc{ExplainaBoard} also considers text generation tasks, and currently mainly focuses on conditional text generation, for example, text summarization~\cite{rush-etal-2015-neural,liu-lapata-2019-text} and machine translation .
System outputs on text summarization are expanded based on the previous work's collection~\cite{bhandari-etal-2020-evaluating} as well as recently state-of-the-art systems \cite{liusimcls2021} while outputs from machine translation are collected from the WMT20.\footnote{\url{http://www.statmt.org/wmt20/metrics-task.html}}

\renewcommand\tabcolsep{2.2pt}
\renewcommand\arraystretch{0.9}  
\begin{table}[htb]
  \centering \footnotesize
    \begin{tabular}{ccccc}
    \toprule
    \multicolumn{2}{c}{\textbf{Task}} & \textbf{Data} & \textbf{Model} &\textbf{Attr.} \\
    \midrule
    \multicolumn{1}{l}{\multirow{3}[6]{*}{Text Classification}} & Sentiment & 8 & 40 &2\\
\cmidrule{2-5}          & Topic & 4 & 18 &2   \\
\cmidrule{2-5}          & Intention & 1  & 3 &2 \\
    \midrule
    \multicolumn{1}{p{6.835em}}{Text-Span Classification} & Aspect Sentiment & 4 & 20 &4   \\
    \midrule
    \multicolumn{1}{p{6.835em}}{Text Pair Classification} & NLI   & 2 & 6 &7    \\
    \midrule
    \multicolumn{1}{l}{\multirow{4}[8]{*}{Sequence Labeling}} & NER   & 3 & 74 &9   \\
\cmidrule{2-5}          & POS   & 3  & 14  &4   \\
\cmidrule{2-5}          & Chunking & 3 & 14  &9    \\
\cmidrule{2-5}          & CWS   & 7 & 64  &7   \\
    \midrule
    \multicolumn{1}{p{6.835em}}{Structure Pred.} & Semantic Parsing & 4 & 12 & 4   \\
        \midrule
    
    \multicolumn{1}{l}{\multirow{2}[3]{*}{Text Generation}} & Summarization & 2 & 36 & 7\\
\cmidrule{2-5}          & Translation & 4 & 60 &9   \\

    \bottomrule
    \end{tabular}%
      \caption{ Brief descriptions of tasks, datasets and systems that \textsc{ExplainaBoard} currently supports. ``{Attr.}'' denotes {Attribute.} ``Pred.'' denotes ``Prediction''.}
  \label{tab:task_data_model}%
\end{table}%

\section{Case Study} \label{sec:case-study}
Here, we briefly showcase the actual \textsc{ExplainaBoard} interface through a case study on analyzing state-of-the-art NER systems.

\subsection{Experimental Setup}

\paragraph{Attribute Definition}
\label{sec:attr-define}
We define attributes following \citet{fu-etal-2020-interpretable} and three of them are used below:
\texttt{entity length}, \texttt{sentence length} and \texttt{label of entity}.

\begin{figure*}%
\centering
\includegraphics[width=0.87\linewidth]{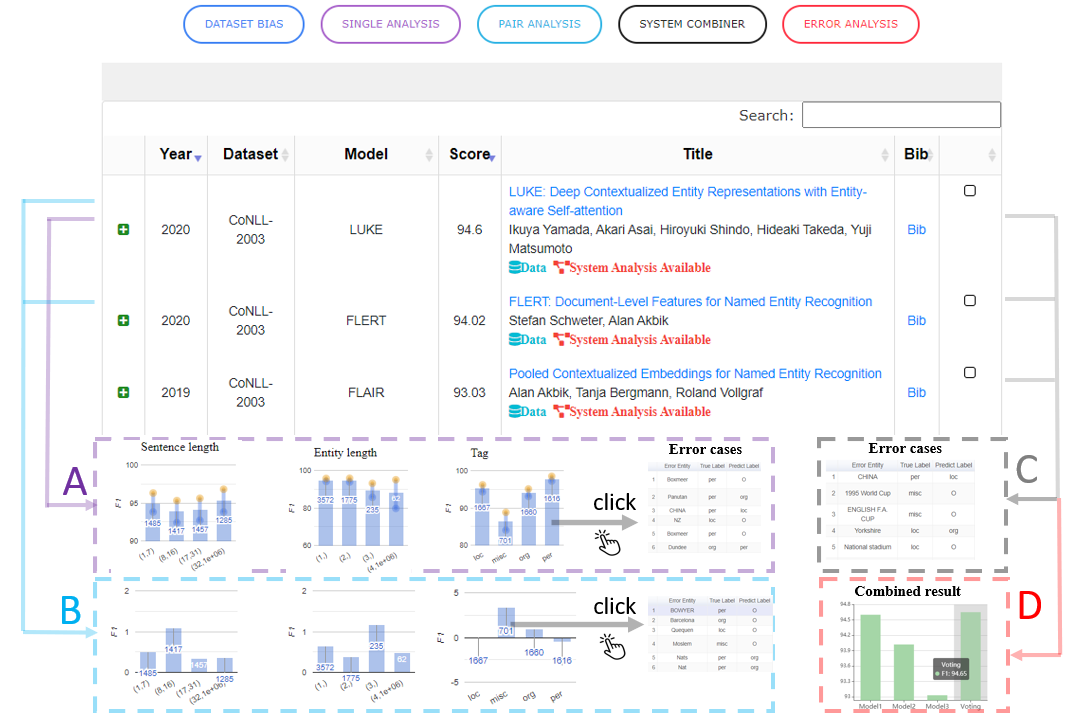}

\caption{
An example of the actual \textsc{ExplainaBoard} interface for NER over three top-performing systems on the \texttt{CoNLL-2003} dataset.
\textbf{Box A} shows the single-system analysis results obtained when users select the top-1 system and click the ``Single Analysis'' button. 
\textbf{Box B} shows the pairwise analysis results when top-2 systems are chosen and ``Pair Analysis'' is clicked. Users can click any bin of the histogram, which results in a fine-grained error case table.
\textbf{Box C} represents a table with common errors of these top-3 systems.
\textbf{Box D} illustrates the combined result of the top-3 systems.
} 
\label{fig:case-study}
\end{figure*}

\paragraph{Collection of Systems Outputs}
Currently, we collect system outputs by either implementing them by ourselves or collecting from other researchers \cite{DBLP:conf/aaai/FuLZ20,schweter2020flert,yamada-etal-2020-luke}.
Using these methods, we have gathered 74 models on six NER datasets with system output information.

\subsection{Analysis using {ExplainaBoard}}

Fig.~\ref{fig:case-study} illustrates different types of results driven by four functionality buttons\footnote{As it is relatively challenging to define calibration in structure prediction tasks, this feature is currently only provided for classification tasks. We will explore more in the future.} over the top-3 NER systems: \texttt{LUKE} \cite{yamada-etal-2020-luke}, \texttt{FLERT} \cite{schweter2020flert} and \texttt{FLAIR}~\cite{akbik-etal-2019-pooled}.

\noindent \textbf{Box A} breaks down the performance of the top-1 system over different attributes.\footnote{Due to the page limitation, we only show three.} We can intuitively observe that even the state-of-the-art system does worse on longer entities. Users can further print error cases in the longer entity bucket by clicking the corresponding bin. 

\noindent \textbf{Box B} shows the 1st system's (\texttt{LUKE}) performance minus the 2nd system's (\texttt{FLERT}) performance. We can see that although \texttt{LUKE} surpasses  \texttt{FLERT} holistically, it performs worse when dealing with \textsc{PERSON} entities.

\noindent \textbf{Box C} identifies samples that all systems mispredict. Further analysis of these samples uncovers challenging patterns or annotation errors.

\noindent \textbf{Box D} examines potential complementarity among these top-3 systems. The result shows that, by a simple voting ensemble strategy, a new state-of-the-art (94.65 F1) has been achieved on the \texttt{CoNLL-2003} dataset.

\section{Usage}

\paragraph{Example Use-cases}
To show the practical utility of \textsc{ExplainaBoard},
we first present examples of how it has been used as an analysis tool in existing published research papers.
\citet{DBLP:conf/aaai/FuLZ20} (Tab.4) utilize  \textit{single-system} analysis with the attribute of \texttt{label consistency} for NER task while \citet{zhong-etal-2019-closer} (Tab.4-5) use it for text summarization with attributes of \texttt{density} and \texttt{compression}.  
 Fig.4 and Tab.3 in \citet{fu-etal-2020-interpretable} leverage the \textit{data bias} analysis and \textit{pairwise system} diagnostics to interpret top-performing NER systems while Tab.4 in \citet{fu-etal-2020-rethinkcws} use \textit{single and pairwise} system analysis to investigate what's next for the Chinese Word Segmentation task.
 \citet{liu2020refactor} use \textit{system combiner} functionality to make ensemble analysis of summarization systems and Fig.1 in \citet{ye21eacl} use \textit{reliability analysis} functionality to observe how confidence intervals change in different buckets of a performance histogram.
 

\paragraph{Using ExplainaBoard}
Researchers can use \textsc{ExplainaBoard} in different ways:
(i) We maintain a website where each task-specific \textsc{ExplainaBoard} allows researchers to interact with it, interpreting systems and datasets that they are interested in from different perspectives.
(ii) We also release our back-end code for different NLP tasks so that researchers could flexibly use them to process their own system outputs, which can assist their research projects.

\paragraph{Contributing to ExplainaBoard}
The community can contribute to \textsc{ExplainaBoard} in several ways:
(i) Submit system outputs of their implemented models.
(ii) Add more informative attributes for different NLP tasks.
(iii) Add new datasets or benchmarks for existing or new tasks.

\section{Implications and Roadmap}

\textsc{ExplainaBoard} presents a new paradigm in leaderboard development for NLP.
This is just the beginning of its development, and there are many future directions.

\paragraph{Research Revolving on System Outputs\footnote{We released system outputs of \textsc{ExplainaBoard}: \url{http://explainaboard.nlpedia.ai/download.html}}}
Due to the ability to analyze, contrast, or combine results from many systems \textsc{ExplainaBoard} incentivizes researchers to submit their results to explainaboard to better understand them and showcase their systems' strengths.
At the same time, \textsc{ExplainaBoard} will serve as a central repository for system outputs across many tasks, allowing for future avenues of research into cross-system analysis or system combination.

\paragraph{Enriching ExplainaBoard with Glass-box Analysis}

\textsc{ExplainaBoard} currently performs black-box analysis, solely analyzing system outputs without accessing model internals.
On the other hand, there are many other glass-box interpretability tools that look at model internals, such as the AllenNLP Interpret~\cite{wallace-etal-2019-allennlp} and Language Interpretability Tool~\cite{tenney2020language}.
Expanding leaderboards to glass-box analysis methods (see \citet{lipton2018mythos,belinkov-glass-2019-analysis} for a survey) is an interesting future work.

In the future, we aim to improve the applicability and usefulness by following action items:
(1) Collaborate with more leaderboard organizers of diverse tasks and set up corresponding \textsc{ExplainaBoard}s for them.
(2) Cover more tasks, datasets, models, as well as functionalities.

\section*{Acknowledgements}
We thanks all reviewers for their valuable comments and authors who share their system outputs with us: 
Ikuya Yamada, Stefan Schweter, Colin Raffel, Yang Liu, Li Dong.
We also thank Vijay Viswanathan, Yiran Chen, Hiroaki Hayashi for useful discussion and feedback about \textsc{ExplainaBoard}.

The work was supported in part by the Air
Force Research Laboratory under agreement number FA8750-19-2-0200. The U.S. Government is authorized to reproduce and distribute reprints
for Governmental purposes notwithstanding any
copyright notation thereon. The views and conclusions contained herein are those of the authors and should not be interpreted as necessarily representing the official policies or endorsements, either expressed or implied, of the Air Force Research Laboratory or the U.S. Government.

\bibliography{anthology,custom}
\bibliographystyle{acl_natbib}



\end{document}